\documentclass[conference]{IEEEtran}
\IEEEoverridecommandlockouts
\usepackage{cite}
\usepackage{amsmath,amssymb,amsfonts}
\usepackage{mathtools}
\usepackage{algorithmic}
\usepackage{graphicx}
\usepackage{textcomp}
\usepackage{xcolor}
\usepackage[thinlines]{easytable}
\usepackage{hyperref}
\usepackage{verbatim}
\usepackage{booktabs}
\usepackage{mathpazo}

\DeclarePairedDelimiter\abs{\lvert}{\rvert}%
\DeclarePairedDelimiter\norm{\lVert}{\rVert}%

\makeatletter
\let\oldabs\abs
\def\abs{\@ifstar{\oldabs}{\oldabs*}}
\let\oldnorm\norm
\def\norm{\@ifstar{\oldnorm}{\oldnorm*}}
\makeatother

\def\BibTeX{{\rm B\kern-.05em{\sc i\kern-.025em b}\kern-.08em
    T\kern-.1667em\lower.7ex\hbox{E}\kern-.125emX}}
\begin{document}

\title{ROS-Neuro Integration of Deep Convolutional Autoencoders for EEG Signal Compression in Real-time BCIs}

\author{\IEEEauthorblockN{Andrea Valenti}
\IEEEauthorblockA{\textit{Dipartimento di Informatica} \\
\textit{Università di Pisa}\\
Pisa, Italy \\
andrea.valenti@phd.unipi.it}
\and
\IEEEauthorblockN{Michele Barsotti}
\IEEEauthorblockA{\textit{Dept. of Biomedical Research} \\
\textit{CAMLIN Italy s.r.l.}\\
Parma, Italy \\
m.barsotti@camlintechnologies.com}
\and
\IEEEauthorblockN{Raffaello Brondi}
\IEEEauthorblockA{\textit{Dept. of Biomedical Research} \\
\textit{CAMLIN Italy s.r.l.}\\
Parma, Italy \\
r.brondi@camlintechnologies.com}\\
\and
\IEEEauthorblockN{Davide Bacciu}
\IEEEauthorblockA{\textit{Dipartimento di Informatica} \\
\textit{Università di Pisa}\\
Pisa, Italy \\
bacciu@di.unipi.it}
\and
\and
\IEEEauthorblockN{}
\IEEEauthorblockA{}
\and
\IEEEauthorblockN{Luca Ascari}
\IEEEauthorblockA{\textit{Dept. of Biomedical Research} \\
\textit{CAMLIN Italy s.r.l.}\\
Parma, Italy \\
l.ascari@camlintechnologies.com}
}

\maketitle

\begin{abstract}
Typical EEG-based BCI applications require the computation of complex functions over the noisy EEG channels to be carried out in an efficient way. Deep learning algorithms are capable of learning flexible nonlinear functions directly from data, and their constant processing latency is perfect for their deployment into online BCI systems. However, it is crucial for the jitter of the processing system to be as low as possible, in order to avoid unpredictable behaviour that can ruin the system's overall usability. In this paper, we present a novel encoding method, based on on deep convolutional autoencoders, that is able to perform efficient compression of the raw EEG inputs. We deploy our model in a ROS-Neuro node, thus making it suitable for the integration in ROS-based BCI and robotic systems in real world scenarios. The experimental results show that our system is capable to generate meaningful compressed encoding preserving to original information contained in the raw input. They also show that the ROS-Neuro node is able to produce such encodings at a steady rate, with minimal jitter. We believe that our system can represent an important step towards the development of an effective BCI processing pipeline fully standardized in ROS-Neuro framework.
\end{abstract}


\section{Introduction}
\label{sec_intro}
In recent years, we have been witnessing an increasing interest in machine learning algorithms applied to Electroencephalography (EEG) signals in the context of Brain-Computer Interfaces (BCI) \cite{2019_zhang_survey}\cite{2019_roy_survey}. EEG-acquisition systems are relatively cheap and portable, and EEG signals can be recorded in a non-invasive way, with very low risk on subjects\cite{2020_gu_survey}. This enables the collection of high amount of EEG data at a low cost.
 \begin{figure}[h]
\begin{center}
\includegraphics[width = 0.45\textwidth]{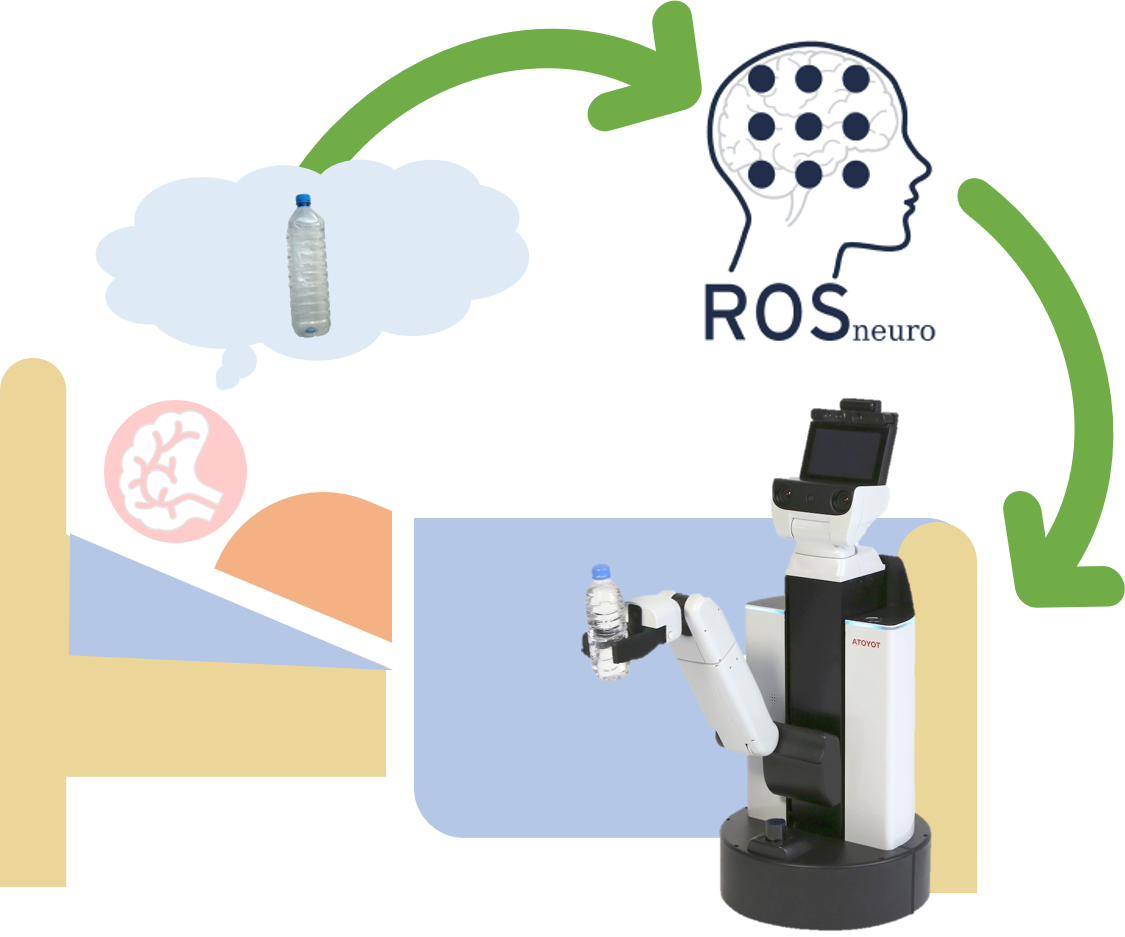}
\end{center}
\caption{Conceptual scheme of the implementation of a  BMI system and a robotic system mutually integrated through ROSNeuro.}
\label{fig:rosneuro_concept}
\end{figure}
Unfortunately, quantity does not entail quality: EEG signals are known to suffer from a low signal-to-noise ratio\cite{2020_gu_survey}, being very sensitive to noise (e.g., electromagnetic). The sources of such noise can be either internal, coming from inside the subject (e.g. eyes blinks or involuntary muscle movements) or external (e.g. power lines and Wi-Fi connections), coming from ambient. Furthermore, the very high temporal resolution of EEG channels is compensated with a low spatial resolution \cite{2020_gu_survey}. In practice, this means that the different EEG channels tends to capture highly-correlated signals. Furthermore, in order to be fully effective, many interesting BCI applications often require very efficient real-time processing of the input signals. 
For instance, consider the case of an assistive-BCI, allowing physically impaired patients to control a robotic arm using his/her EEG signals. In order for the system to be useful, the motion commands need to be acquired, decoded, processed and sent without the presence of noticeable delays which can hamper the overall performance of the system making it unusable for real-world scenarios. Furthermore, the lack of a standardized interface between BCI systems and robotic systems, often leads to the development of very specific solution featuring a limited portability and a restricted exploitation of the robotic devices.
In this scenario, the recent open-source framework ROS-Neuro (formerly known as ROS-health \cite{2018_Beraldo_ROS_health}), comes to enhance the interaction between the worlds of BCI and robotics, by relying on the middleware that has become the standard de-facto for robotic applications: Robot Operating System (ROS)\footnote{\texttt{\url{https://www.ros.org/}}}. 
Indeed, several similarities between the Robot Operating System (ROS) and BCI systems exist: both of them are based on processes running in parallel and communicating each other with the minimum possible delay. In the ROS framework, such modules (nodes) are in charge of different  aspects of the robotic applications. Thus, the ROS communication infrastructure perfectly matches the requirement of any BCI system. For example, a first approach  for integrating ROS and BCI can be found in Beraldo et al.\cite{2018_Beraldo_BCImeetROS}, with the goal of mentally drive a telepresence robot.
Furthermore, the modularity of the architecture and the reliability of the communication infrastructure provide the ROS-Neuro framework with the double fold advantage of integrating both the BCI loop and the robotic control in the same ecosystem and of being supported and developed by a constantly growing community.
This allows researchers to have concurrent access to the user’s neural signals and to the robot (see, for example, the work of Tonin et al. \cite{2019_Tonin_ROS_neuro}), thus effectively enabling the development of new interaction solutions for BCI actuated devices. \\

As deep learning algorithms (DL) have already been proven capable of successfully processing physiological data \cite{2019_roy_survey}\cite{2019_zhang_survey}, it follows that the natural next step is the integration of DL models in the ROS-Neuro framework.
In this paper, we argue that DL models and ROS-Neuro can be effectively combined for real-time BCI task and we present for the first time a ROS-Neuro node capable to perform efficient real-time encoding of EEG-based BCI signals. 
Specifically, we employ a deep autoencoder (AE) in order to perform dimensionality reduction of the raw EEG channels in real-time. This allow us to obtain a cleaned and more compact representation of the input, thus boosting the processing time of any subsequent module of the system in terms of booth speed and accuracy. 
The rest of this paper is organized as follows: in Sec. \ref{sec_related_work}, we report some applications of AE-based DL models for BCI tasks. In Sec. \ref{sec_methods}, we present our system. First, we accurately describe the AE model, then we show how such model can be deployed on a ROS-Neuro node. In Sec. \ref{sec_results} we present the quantitative result we obtained on the signal compression task, showing that our model is able to compress the raw input signal into an efficient representation that can be easily used by downstream nodes for further processing, with a very limited time jitter. Finally, in Sec. \ref{sec_conclusion} we summarize the main points of this paper and provide some possible future developments of this work.

\section{Autoencoder-based deep learning for BCI}
\label{sec_related_work}
In the early years, researchers were mostly focusing on enhancing the performance of traditional non-deep classifier with several kinds of AE, pre-trained on the input in an unsupervised way before the actual classifier. For example, in one of the early works on machine learning and BCI \cite{2014_Li_Multifractal}, Li et al. use a combination of deep belief networks and denoising AE to develop a classifier able to reach up to 90\% accuracy on some motor imagery tasks. They also tackle the problem of classification of motor imagery data when some part of the signals have been removed due to artefacts corruption \cite{2014_Li_Feature_Learning}. They do so by extracting frequency information from the incomplete input using Lomb-Scargle periodgrams, which is then fed to a classifier composed of a combination of denoising AE and SVM.
One of the main challenges of BCI systems is to achieve good performance across different subjects. Stober et al. \cite{2016_Stober_Deep_Feature} explore different methods for enforcing subject-invariant representations. Using cross-trial encodings and similarity constraint encodings, they are able to achieve superior inter-subject classification accuracy than standard methods on a challenging dataset composed of EEG signals of subjects listening to music. Zhang et al. \cite{2017_Zhang_Multi-Person} try to solve the inter-subject classification problem by combining an AE for feature learning and gradient-boosted decision trees for a motor imagery classification task, achievening almost 75\% of accuracy on multiple subjects.
Deep learning models have been also applied with success to the P300 component recognition task. Vařeka et al.\cite{2017_Vareka_Stacked} develop a sparse AE for P300 recognition that is able to outperform similar machine learning classifiers. Another interesting work is the one by \cite{2019_Ditthapron_Universal} Ditthapron et al., where they detect the P300 component from raw EEG signals. Input data is arranged in a 2D matrix in order to preserve the spatial dependencies between the electrodes. The model uses a combination of convolutional layers for the extraction of spatial features and recurrent layers for learning the temporal relationships of the signals.
In more recent years, the rapid increase of computing power, along with some key advances in the field, has allowed the use of bigger and deeper models, jointly trained end-to-end. For example, an interesting line of work is the one by Tan et al. \cite{2019_Tan_Attention}, which use a convolutional, attention-based model for the classification of listened short musical fragments. The model is trained on EEG optical flow data, and an adversarial network is used to force the encoder to extract domain-invariant representations. Thus, they are able to reach around 37\% accuracy on the task. The idea of using adversarial networks to achieve representation invariance is also used by Özdenizci et al. \cite{2019_Ozdenici_Tansfer}. They are able to obtain more than 60\% of accuracy on all subjects, outperforming variants that do not leverage the adversarial network. Interesting is the work by Lee et al. \cite{2020_Lee_DecodingMI}, that leverage motor-execution EEG data for the classification of motor imagery task using a relation network \cite{2017_Santoro_asimple}. They can obtain superior results than similar models that use just one type of data, either ME or MI. An example of deep CNN for a motor preparation task can be found in \cite{2020_Mammone_DeepCNN}, where Mammone et al. use beamforming to solve the inverse problem of motor cortex source reconstruction. They then feed the preprocessed data into a CNN to perform state of the art classification of motor preparation tasks.
For extended surveys on machine learning models applied on EEG-based signals, we refer the interested reader to \cite{2019_zhang_survey} and \cite{2019_roy_survey}.


\section{Methods}
\label{sec_methods}
\subsection{Dataset}
\label{sec_dataset}
The dataset used in this work is freely available at the BNCI Horizon 2020 website\footnote{\texttt{\url{http://bnci-horizon-2020.eu/database/data-sets}}} and described in detail in \cite{2017_Ofner_Upper_Limb}. 
It consists in both motor execution and motor imagery EEG data from a series of 61 electrodes, distributed over the frontal, central, parietal and temporal areas of the skull. EEG data is recorded along with motion data information, coming from both a glove and a exoskeleton. The data acquisition frequency is 512Hz and the dataset was recorded using active electrodes and four 16-channel amplifiers. The dataset involves 15 healthy subjects (aged between 22 and 40 years), 9 female and 6 males. The subjects are required to perform six types of actions (elbow flection, elbow extension, wrist pronation, wrist supination, hand open, hand close), plus an additional ``rest action", for a total of 7 different movement classes. Ten runs of data acquisitions are recorded for each subject, each run containing 42 different trials, equally divided between the 7 classes.
Each trial is structured as follows: at second 0, a beep sounded and a green cross was displayed on computer screen. Subjects were instructed to focus their gaze to the cross. At second 2, a cue was displayed on the screen, showing one of the movements to perform. The subjects were required to perform the specified movement for the next 3 seconds, then go back to the resting position. After the movement, subjects waited in the resting position for a random amount of time, (between 2 and 3 seconds), before the beginning of the next trial.

\subsection{Data Preprocessing}

\begin{table}
\caption{Spatial arrangement of input channels for each timestep.}
\begin{center}
\begin{TAB}(e,0.5cm,0.5cm){|c:c:c:c:c:c:c:c:c|}{|c:c:c:c:c:c:c:c:c:c|}
     &      & F3   & F1   & Fz  & F2   & F4   &      &      \\
     & FFC5 & FFC3 & FFC1 &     & FFC2 & FFC4 & FFC6 &      \\
     & FC5  & FC3  & FC1  & FCz & FC2  & FC4  & FC6  &      \\
FTT7 & FCC5 & FCC3 & FCC1 &     & FCC2 & FCC4 & FCC6 & FTT8 \\
     & C5   & C3   & C1   & Cz  & C2   & C4   & C6   &      \\
TTP7 & CCP5 & CCP3 & CCP1 &     & CCP2 & CCP4 & CCP6 & TTP8 \\
     & CP5  & CP3  & CP1  & CPz & CP2  & CP4  & CP6  &      \\
     & CPP5 & CPP3 & CPP1 &     & CPP2 & CPP4 & CPP6 &      \\
     &      & P3   & P1   & Pz  & P2   & P4   &      &      \\
     &      &      & PPO1 &     & PPO2 &      &      &      \\
\end{TAB}
\end{center}
\label{tab_channel_map}
\end{table}

\label{sec_preprocessing}
Firstly, in order to remove power line noise, a notch filter at 50Hz is applied. Then, the EEG signals are bandpass filtered in the range 0.5Hz - 60Hz using a 5th-order Butterworth filter and down-sampled to 128Hz. 
We excluded Subject 1 from the training data because channel names were not available for this subject, making impossible to perform the remaining preprocessing steps, as described in the rest of this section.
It is well known that the spatial information of the EEG plays a crucial role for the analysis of the signals and that the information associated with each electrode is dependent on its specific position on the skull surface.


In order to make the raw EEG channels suitable to be processed by a CNN, we arrange them on a 2D $10\times9$ matrix as shown in Tab. \ref{tab_channel_map}. Elements of the matrix that do not correspond to any channel are set to 0. We decide to encode chunks of 16 timesteps (corresponding to 1/8 of a second for signals sampled at 128Hz) into a single latent representation, hence effectively turning each data sample into a 3D matrix $ x \in \mathbb{R}^{16\times10\times9} $ . 
This additional processing steps, which can be done at virtually no cost, is performed in order to make the DL algorithm able to capture  local features that depends on groups of neighbouring channels instead of just a single channel, resulting in more expressive learned representations than the one learned by a simple multi-layer perceptron. 

\subsection{Deep Autoencoder}
\label{sec:deep_autoencoder}
The input representation described in Sec. \ref{sec_preprocessing}, albeit powerful, is not the optimal choice as it is susceptible to the \emph{curse of dimensionality} (i.e. a rapid decrease of performance/increase of training time as the dimensionality of the input becomes larger), a well-known problem for many machine learning classifiers of common use.
From a spatial point of view, many elements of the input matrix are set to 0, thus conveying no information for the downstream task. In addition, the remaining channels tends to carry highly correlated information, making such a high number of input redundant. From a temporal point of view, the sampling frequency is generally unnecessarily high, since most of the interesting information of EEG is contained into the relatively slow temporal evolution of the channels rather than in the specific local nuances. Therefore, it is useful to encode a certain portion of the input signal into a compact representation that still retains all the information of the original input (and, possibly, discards some of the noise).
The AE model is an unsupervised deep learning model composed by two main sub-models: an encoder $z = f(x)$ that learns to transform an input $x \in \mathbb{R}^{N_x}$ into a compact latent code $z \in \mathbb{R}^{N_z}$, and a decoder $\tilde{x} = g(z)$ that learns the inverse transformation, trying to reconstruct the original $x$ from $z$. The AE is trained end-to-end through gradient descent to reconstruct its input according to the following loss function:
\begin{equation}
\label{eq_mse}
    \mathcal{L}_{AE} = \norm{x - \tilde{x}}_2^2
\end{equation}

In the typical setting, we have $N_z < N_x$, so that at the end of training the latent code $z$ will contain all the necessary information needed to reconstruct $x$. $z$ is thus a compressed representation of $x$. After training, the encoder can then be used to perform efficient (constant-time) compression of the input.
The architecture of our AE is composed as follows: the input, shaped in its 3D representation as described in Sect. \ref{sec_preprocessing}, is fed to two 3-D convolutional layers of 16 and 32 neurons respectively, with ReLU activation functions. Each convolutional layer is followed by a batch normalization layer \cite{2015_ioffe_batchnorm}. After the convolutions, we perform max-pooling of the resulting 3D channels in order to increase the local spatio-temporal invariance of the obtained representation. We then flatten the result into a 1D vector, and feed it into a fully connected layer of 128 neurons, thus obtaining the final code $z$. The decoder's architecture is specular to the encoder's, using 3D deconvolutions (also known as transposed convolutions) instead of convolutions, and a max-unpooling layer to reverse the effects of the max pooling. Note that, starting from a 3D matrix of $16*10*9=1440$ dimensions, we reach a final code dimension of 128, thus achieving a compression ratio of $11.25$. In other words, the final size of $z$ is more than $90\%$ smaller than the initial size. This can greatly speed up the computation of downstream nodes of the system, making this model particularly suited for task that requires online data processing. 
The implemented network architecture has been empirically defined and the model selection (in terms of hyperparameters such as learning rate, number and size of layers,  dimensionality of the latent code) have been performed through a 5-fold cross validation.

\subsection{ROS-Neuro Node}
\label{sec_rosnode}

The current version of ROS-Neuro \cite{2019_Tonin_ROS_neuro} comes with three main ROS packages:

\begin{itemize}
    \item The \verb+rosneuro_msgs+ package defines ROS messages describing data formats to be exchanged between nodes.
    \item The \verb+rosneuro_acquisition+ package contains a ROS node called \verb+Acquisition+ representing the main entry point of the ROSNeuro graph.
    \item The \verb+rosneuro_recorder+ package allows to store messages circulating on the \verb+/neurodata+ topic on GDF or BDF files.
\end{itemize}

\begin{figure}[t]
\begin{center}
\includegraphics[width = 0.5\textwidth]{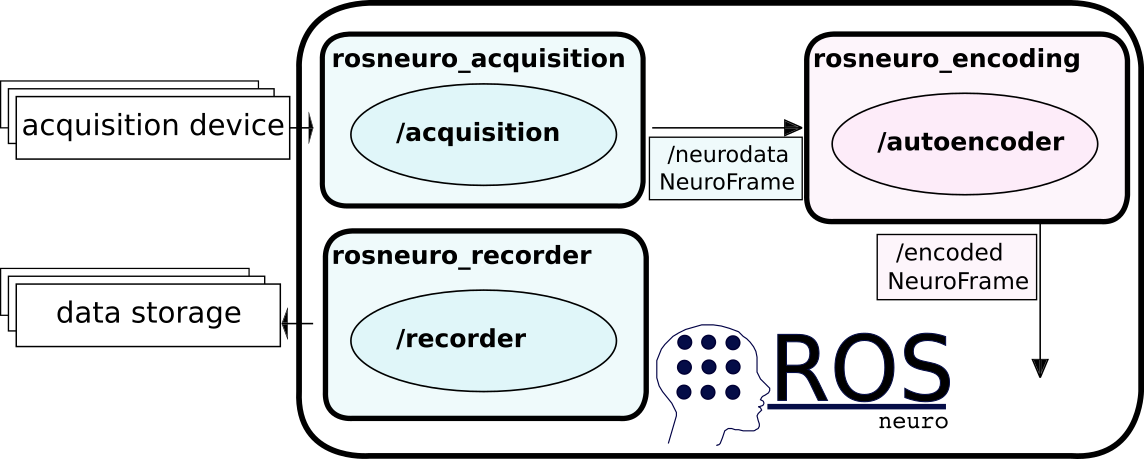}
\end{center}
\caption{Schematic representation of the ROS-Neuro framework. Blue colored boxes represent currently available packages whereas the package developed in this work is colored in red.}
\label{fig:schematicROS}
\end{figure}

The \verb+rosneuro_acquisition+ package contains a ROS node called \verb+Acquisition+ representing the main entry point of the ROS-Neuro graph.
The node uses the library \verb+libeegdev+\footnote{\texttt{\url{http://neuro.debian.net/pkgs/libeegdev-dev.html}}} in order to  to interface with several different commercial acquisition devices and then convert them in \verb+NeuroFrames+ messages sent through the \verb+/neurodata+ topic. The \verb+NeuroFrame+ message defines the main input of the ROS-Neuro graph incapsulating readings from an amplifiers. Moreover, \verb+Acquisition+ node is also able to simulate an ongoing stream by the playing-back of a GDF format file (General Data Format for biomedical signals).
In this section we present a new ROS-Neuro node, called \verb+autoencoder+ and implemeted in the \verb+rosneuro_encoding+ package with the aim of reducing the noise in the EEG signals and providing a more compact and meaningful representation of the data.
The \verb+autoencoder+ node, written in python and implemented using \verb+rospy+, subscribes to the \verb+/neurodata+ topic published by the acquisition node (see Figure \ref{fig:schematicROS}) receiving then a frame of EEG data (the size of the published frame depends on the original sampling frequency and on the given frame rate which is set equal to 16Hz resulting in 32 samples at 512Hz).
The node than performs the processing explained in section \ref{sec:deep_autoencoder}. In particular, the offline trained AE model is loaded using the PyTorch\footnote{\texttt{\url{https://pytorch.org/}}} library at then, for each frame, the node maps the incoming signals into the matrix depicted in Figure \ref{tab_channel_map}. Finally, the output of the AE model is published on a new topic, called \verb+/encoded+, in form of NeuroFrame message. The time interval between two consecutive outputs of the node has been taken as an index of stability of the system.

\begin{table}[t]
    \centering
    \caption{Mean squared error of model's reconstructions on the test set. Mean and standard deviation are computed across the different folds.}
    \begin{tabular}{lcc}
    \toprule
    Subject    &  MSE (mean, $\mu V$) & MSE (std, $\mu V$) \\
    \midrule
    S02         &   0.027     &   0.0005         \\
    S03         &   0.552     &   0.1360         \\
    S04         &   0.172     &   0.0150         \\
    S05         &   0.023     &   0.0011         \\
    S06         &   0.028     &   2.4 $\times 10^{-5}$         \\
    S07         &   0.019     &   0.0001         \\
    S08         &   0.032     &   0.0009         \\
    S09         &   0.026     &   0.0001         \\
    S10         &   0.016     &   0.0004         \\
    S11         &   0.041     &   0.1 $\times 10^{-8}$        \\
    S12         &   0.018     &   8.6 $\times 10^{-5}$       \\
    S13         &   0.018     &   0.0188         \\
    S14         &   0.078     &   0.0013         \\
    S15         &   0.036     &   0.0164         \\
    \bottomrule
    \end{tabular}
    \label{tab_mse}
\end{table}


\section{Results}
\label{sec_results}
In order to validate our approach, we firstly investigate how well the implemented deep AE model is able to reconstruct its input. The ability of our model to perform accurate reconstruction is crucial, since we do not want to discard potentially useful information at this stage of the processing pipeline. For this purpose, we consider the dataset previously described in Sec. \ref{sec_dataset}, preprocessed according to the procedure described in Sec. \ref{sec_preprocessing}. We evaluate the performance of our model using a 3-fold cross-validation approach, repeated 3 times with different train/test splits. Thus, at each fold, the model is trained on about  67\% of the dataset and tested on the remaining 33\%. The model was trained and tested on each subject individually. 
The training is performed on a single NVIDIA Tesla V100 GPU. The final reconstruction mean squared error (MSE), measuring the absolute difference between the original data samples and their reconstructions, is computed on the test set according to Eq. \ref{eq_mse} and is reported on Tab. \ref{tab_mse}. 
The results show that the model is able to successfully reconstruct its input with very low error, meaning that the compressed code $z$ is an informative representation of the input. The model's performance are stable, with low standard deviation between the different folds. Even the MSE across the subjects is generally similar. The slightly higher reconstruction errors of subjects 3 and 4 are due to a higher presence of noise (mostly electrode artifacts and muscular artifacts) with respect to the EEG signals of other subjects. Notably, these are the subjects that allowed to obtain the highest performances in a previous study on the same dataset\cite{2020_Mammone_DeepCNN}. It will be interesting, in future work, to explore whether the proposed encoding approach is effective in maintaining such high accuracy.
These are further signs that our model's architecture is expressive enough to capture the essential information contained in the inputs. Using a CNN increases the spatio-temporal invariance of the representations, making the model more robust to slightly different placements of the EEG acquisitions device, as well as the different reaction times of the subjects.

\begin{figure}[t]
\begin{center}
\includegraphics[width =0.45\textwidth]{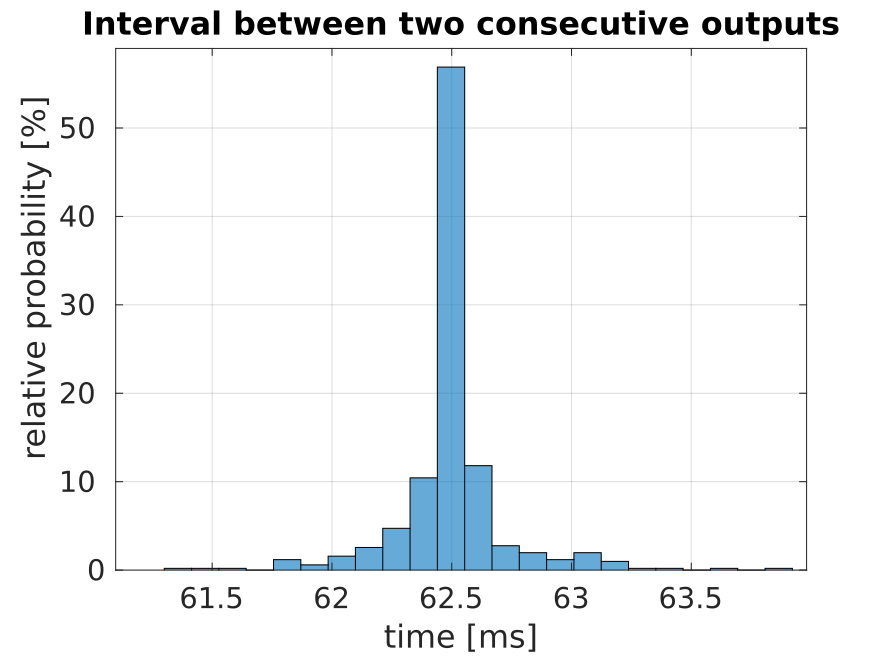}
\end{center}
\caption{Relative frequency histogram of the distribution of intervals between two consecutive outputs of the $rosneuro\_encoding$  $autoencoder$ node. The sum of the bar weight is 100\%.}
\label{fig_jitter}
\end{figure}

Since our goal is to to deploy the system for real-time online tasks, we investigated the stability of the implemented ROS-Neuro node, i.e. whether the predictions of the trained AE running inside the implemented ROS-Neuro node are affected by delays. We would like the model to be able to process the input stream at a constant rate, thus avoiding processing bottlenecks and unpredictable behaviour. In order to check out which are the performance in terms of stability, we examined the time interval between two consecutive messages in the topic \verb+/encoded+. Since the frame rate of the \verb+/acquisition+ node is set equal to 16Hz, the interval between two output is accessible every 62.5 ms. In Figure \ref{fig_jitter}, the normalized histogram of the collected time interval between two consecutive output of the presented node is reported. It can be noted that more than 50\% of the samples are exactly in the expected range and that the system is highly stable in terms of latency featuring an average time elapsed of $62.499 \pm 0.227 ms$.

\section{Conclusion}
\label{sec_conclusion}
In this paper, we described a novel deep Autoencoder architecture, that is capable of performing real-time compression of input data via its deployment inside a ROS-Neuro node. We showed how deep learning and the ROS-Neuro system can be effectively combined in order to efficiently perform complex functions on an real-time incoming stream of EEG signals. The empirical evidence showed that our model is able to achieve, with an almost constant latency and very low jitter, a very high compression ratio of the input signal without losing important information.
Although the results confirm our initial hypothesis, this line of work is still at its initial stages, and room for improvements exist. It also may be useful to explore a more refined version of the EEG channels's layout, leveraging the exact spatial coordinates of the inputs, which can then be converted to a point cloud and processed using a geometric neural network \cite{2019_spurek_geometric}. Our works pave the way for exploring additional (possibly deep) classifiers to the output of the ROS-Neuro node. The architecture of the AE can be further refined to be more robust to specific type of artifacts that can be commonly found in raw EEG signals, such as eye movements. 

\section*{Acknowledgment}
The authors wish to thank Toyota Motor Europe and Toyota Motor Corporation for providing CAMLIN with the Human Support Robot (HSR), the robotic platform around which CAMLIN is developing its BCI for assisted living program.
\bibliographystyle{IEEEtran}
\bibliography{IEEEabrv,smc2020}

\begin{thebibliography}{10}
\providecommand{\url}[1]{#1}
\csname url@samestyle\endcsname
\providecommand{\newblock}{\relax}
\providecommand{\bibinfo}[2]{#2}
\providecommand{\BIBentrySTDinterwordspacing}{\spaceskip=0pt\relax}
\providecommand{\BIBentryALTinterwordstretchfactor}{4}
\providecommand{\BIBentryALTinterwordspacing}{\spaceskip=\fontdimen2\font plus
\BIBentryALTinterwordstretchfactor\fontdimen3\font minus
  \fontdimen4\font\relax}
\providecommand{\BIBforeignlanguage}[2]{{%
\expandafter\ifx\csname l@#1\endcsname\relax
\typeout{** WARNING: IEEEtran.bst: No hyphenation pattern has been}%
\typeout{** loaded for the language `#1'. Using the pattern for}%
\typeout{** the default language instead.}%
\else
\language=\csname l@#1\endcsname
\fi
#2}}
\providecommand{\BIBdecl}{\relax}
\BIBdecl

\bibitem{2019_zhang_survey}
X.~Zhang, L.~Yao, X.~Wang, J.~Monaghan, D.~Mcalpine, and Y.~Zhang, ``A survey
  on deep learning based brain computer interface: Recent advances and new
  frontiers,'' 2019.

\bibitem{2019_roy_survey}
Y.~Roy, H.~Banville, I.~M. Carneiro~de Albuquerque, A.~Gramfort, T.~Falk, and
  J.~Faubert, ``Deep learning-based electroencephalography analysis: a
  systematic review,'' \emph{Journal of Neural Engineering}, vol.~16, 05 2019.

\bibitem{2020_gu_survey}
X.~Gu, Z.~Cao, A.~Jolfaei, P.~Xu, D.~Wu, T.-P. Jung, and C.-T. Lin, ``Eeg-based
  brain-computer interfaces (bcis): A survey of recent studies on signal
  sensing technologies and computational intelligence approaches and their
  applications,'' 2020.

\bibitem{2018_Beraldo_ROS_health}
G.~Beraldo, N.~Castaman, R.~Bortoletto, E.~Pagello, J.~d.~R. Millan, L.~Tonin,
  and E.~Menegatti, ``Ros-health: An open-source framework for neurorobotics,''
  05 2018.

\bibitem{2018_Beraldo_BCImeetROS}
G.~{Beraldo}, M.~{Antonello}, A.~{Cimolato}, E.~{Menegatti}, and L.~{Tonin},
  ``Brain-computer interface meets ros: A robotic approach to mentally drive
  telepresence robots,'' in \emph{2018 IEEE International Conference on
  Robotics and Automation (ICRA)}, 2018, pp. 4459--4464.

\bibitem{2019_Tonin_ROS_neuro}
L.~{Tonin}, G.~{Beraldo}, S.~{Tortora}, L.~{Tagliapietra}, J.~d.~R.~{Millán},
  and E.~{Menegatti}, ``Ros-neuro: A common middleware for bmi and robotics.
  the acquisition and recorder packages,'' in \emph{2019 IEEE International
  Conference on Systems, Man and Cybernetics (SMC)}, 2019, pp. 2767--2772.

\bibitem{2014_Li_Multifractal}
J.~Li and A.~Cichocki, ``Deep learning of multifractal attributes from motor
  imagery induced eeg,'' in \emph{Neural Information Processing}, C.~K. Loo,
  K.~S. Yap, K.~W. Wong, A.~Teoh, and K.~Huang, Eds.\hskip 1em plus 0.5em minus
  0.4em\relax Cham: Springer International Publishing, 2014, pp. 503--510.

\bibitem{2014_Li_Feature_Learning}
\BIBentryALTinterwordspacing
J.~Li, Z.~Struzik, L.~Zhang, and A.~Cichocki, ``Feature learning from
  incomplete eeg with denoising autoencoder,'' \emph{Neurocomputing}, vol. 165,
  pp. 23 -- 31, 2015. [Online]. Available:
  \url{http://www.sciencedirect.com/science/article/pii/S0925231215004282}
\BIBentrySTDinterwordspacing

\bibitem{2016_Stober_Deep_Feature}
\BIBentryALTinterwordspacing
S.~Stober, A.~Sternin, A.~M. Owen, and J.~A. Grahn, ``Deep feature learning for
  {EEG} recordings,'' \emph{CoRR}, vol. abs/1511.04306, 2015. [Online].
  Available: \url{http://arxiv.org/abs/1511.04306}
\BIBentrySTDinterwordspacing

\bibitem{2017_Zhang_Multi-Person}
\BIBentryALTinterwordspacing
X.~Zhang, L.~Yao, D.~Zhang, X.~Wang, Q.~Z. Sheng, and T.~Gu, ``Multi-person
  brain activity recognition via comprehensive {EEG} signal analysis,''
  \emph{CoRR}, vol. abs/1709.09077, 2017. [Online]. Available:
  \url{http://arxiv.org/abs/1709.09077}
\BIBentrySTDinterwordspacing

\bibitem{2017_Vareka_Stacked}
\BIBentryALTinterwordspacing
L.~Vařeka and P.~Mautner, ``Stacked autoencoders for the p300 component
  detection,'' \emph{Frontiers in Neuroscience}, vol.~11, p. 302, 2017.
  [Online]. Available:
  \url{https://www.frontiersin.org/article/10.3389/fnins.2017.00302}
\BIBentrySTDinterwordspacing

\bibitem{2019_Ditthapron_Universal}
A.~Ditthapron, N.~Banluesombatkul, S.~Ketrat, E.~Chuangsuwanich, and
  T.~Wilaiprasitporn, ``Universal joint feature extraction for p300 eeg
  classification using multi-task autoencoder,'' \emph{IEEE Access}, vol.~PP,
  pp. 1--1, 05 2019.

\bibitem{2019_Tan_Attention}
C.~{Tan}, F.~{Sun}, T.~{Kong}, B.~{Fang}, and W.~{Zhang}, ``Attention-based
  transfer learning for brain-computer interface,'' in \emph{ICASSP 2019 - 2019
  IEEE International Conference on Acoustics, Speech and Signal Processing
  (ICASSP)}, 2019, pp. 1154--1158.

\bibitem{2019_Ozdenici_Tansfer}
O.~{Özdenizci}, Y.~{Wang}, T.~{Koike-Akino}, and D.~{Erdoğmuş}, ``Transfer
  learning in brain-computer interfaces with adversarial variational
  autoencoders,'' in \emph{2019 9th International IEEE/EMBS Conference on
  Neural Engineering (NER)}, 2019, pp. 207--210.

\bibitem{2020_Lee_DecodingMI}
D.-Y. Lee, J.-H. Jeong, K.-H. Shim, and S.~Lee, ``Decoding movement imagination
  and execution from eeg signals using bci-transfer learning method based on
  relation network.'' \emph{arXiv: Signal Processing}, 2020.

\bibitem{2017_Santoro_asimple}
A.~Santoro, D.~Raposo, D.~G.~T. Barrett, M.~Malinowski, R.~Pascanu, P.~W.
  Battaglia, and T.~P. Lillicrap, ``A simple neural network module for
  relational reasoning,'' in \emph{NIPS}, 2017.

\bibitem{2020_Mammone_DeepCNN}
\BIBentryALTinterwordspacing
N.~Mammone, C.~Ieracitano, and F.~C. Morabito, ``A deep cnn approach to decode
  motor preparation of upper limbs from time–frequency maps of eeg signals at
  source level,'' \emph{Neural Networks}, vol. 124, pp. 357 -- 372, 2020.
  [Online]. Available:
  \url{http://www.sciencedirect.com/science/article/pii/S089360802030037X}
\BIBentrySTDinterwordspacing

\bibitem{2017_Ofner_Upper_Limb}
\BIBentryALTinterwordspacing
P.~Ofner, A.~Schwarz, J.~Pereira, and G.~R. Müller-Putz, ``Upper limb
  movements can be decoded from the time-domain of low-frequency eeg,''
  \emph{PLOS ONE}, vol.~12, no.~8, pp. 1--24, 08 2017. [Online]. Available:
  \url{https://doi.org/10.1371/journal.pone.0182578}
\BIBentrySTDinterwordspacing

\bibitem{2015_ioffe_batchnorm}
S.~Ioffe and C.~Szegedy, ``Batch normalization: Accelerating deep network
  training by reducing internal covariate shift,'' in \emph{Proceedings of the
  32nd International Conference on International Conference on Machine Learning
  - Volume 37}, ser. ICML’15.\hskip 1em plus 0.5em minus 0.4em\relax
  JMLR.org, 2015, p. 448–456.

\bibitem{2019_spurek_geometric}
P.~Spurek, T.~Danel, J.~Tabor, M.~Śmieja, Łukasz Struski, A.~Słowik, and
  Łukasz Maziarka, ``Geometric graph convolutional neural networks,'' 2019.

\end{thebibliography}

\end{document}